\documentclass[10pt,twocolumn,letterpaper]{article}

\usepackage{cvpr}              

\usepackage[dvipsnames]{xcolor}

\definecolor{cvprblue}{rgb}{0.21,0.49,0.74}
\usepackage[pagebackref,breaklinks,colorlinks,citecolor=cvprblue]{hyperref}

\usepackage{booktabs}
\usepackage{multirow}
\usepackage{comment}

\def\paperID{*****} 
\def\confName{CVPR}

\def\ours{InfMLLM}

\title{InfMLLM: A Unified Framework for Visual-Language Tasks}

\makeatletter
\def\thanks#1{\protected@xdef\@thanks{\@thanks
        \protect\footnotetext{#1}}}
\makeatother
\author{
    \text{Qiang Zhou$^\dag$ \quad Zhibin Wang$^\dag$ \quad Wei Chu \quad Yinghui Xu \quad Hao Li$^*$ \quad Yuan Qi$^*$} \\
    \texttt {\{zhouqiang, \thinspace zhibin.waz, \thinspace chuwei, \thinspace xuyinghui, \thinspace qiyuan\}@inftech.ai} \\
    \texttt {lihao\_hank@163.com}
    \thanks{$^\dag$ Equal contribution.}
    \thanks{$^*$ Corresponding author.}
}

\begin{document}
\maketitle

\begin{abstract}

Large language models (LLMs) have proven their remarkable versatility in handling a comprehensive range of language-centric applications. To expand LLMs' capabilities to a broader spectrum of modal inputs, multimodal large language models (MLLMs) have attracted growing interest. 
This work delves into enabling LLMs to tackle more vision-language-related tasks, particularly image captioning, visual question answering (VQA,) and visual grounding. 
To this end, we implemented a three-stage training scheme: starting with lightweight alignment pretraining, then moderate-weight multitask hybrid training, and finally, LLM fine-tuning to improve instruction following capability. Throughout the training process, the requirements on GPU memory gradually increase.
To effectively manage the number of visual embeddings passed to the LLM while preserving their positional information, we introduce a straightforward visual adapter module dubbed pool-adapter. Our experiments demonstrate that preserving the positional information of visual embeddings through the pool-adapter is particularly beneficial for tasks like visual grounding.
We name our proposed approach InfMLLM and have evaluated it extensively on various benchmark datasets. Our results demonstrate that InfMLLM achieves either state-of-the-art (SOTA) performance or performance comparable to recent MLLMs.
The code and model will be made open-source at: \url{https://github.com/infly-ai/INF-MLLM}.

\end{abstract}    
\section{Introduction}
\label{sec:intro}

As interest in large language models (LLMs) grows, so does the research community's focus on multimodal large language models (MLLMs). MLLMs are essential in developing versatile general-purpose assistants, as everyday interactions involve information of various modalities, such as voice, text, images, and videos. This work extends LLMs to handle more vision-language-related tasks, including image captioning, VQA, and visual grounding. 
Developing an MLLM from scratch is a challenging task. It necessitates substantial training resources and high-quality training data in large quantities. Recent MLLMs~\cite{li2023blip2,instructblip,llava_v15_liu2023improved,chen2023minigptv2,Qwen-VL} instead utilize finetuning strategies, which leverage pretrained LLMs and finetuning them using multimodal data.

\begin{figure}[!t]
    \centering
    \begin{subfigure}[b]{0.92\linewidth}
         \centering
         \includegraphics[width=0.92\linewidth]{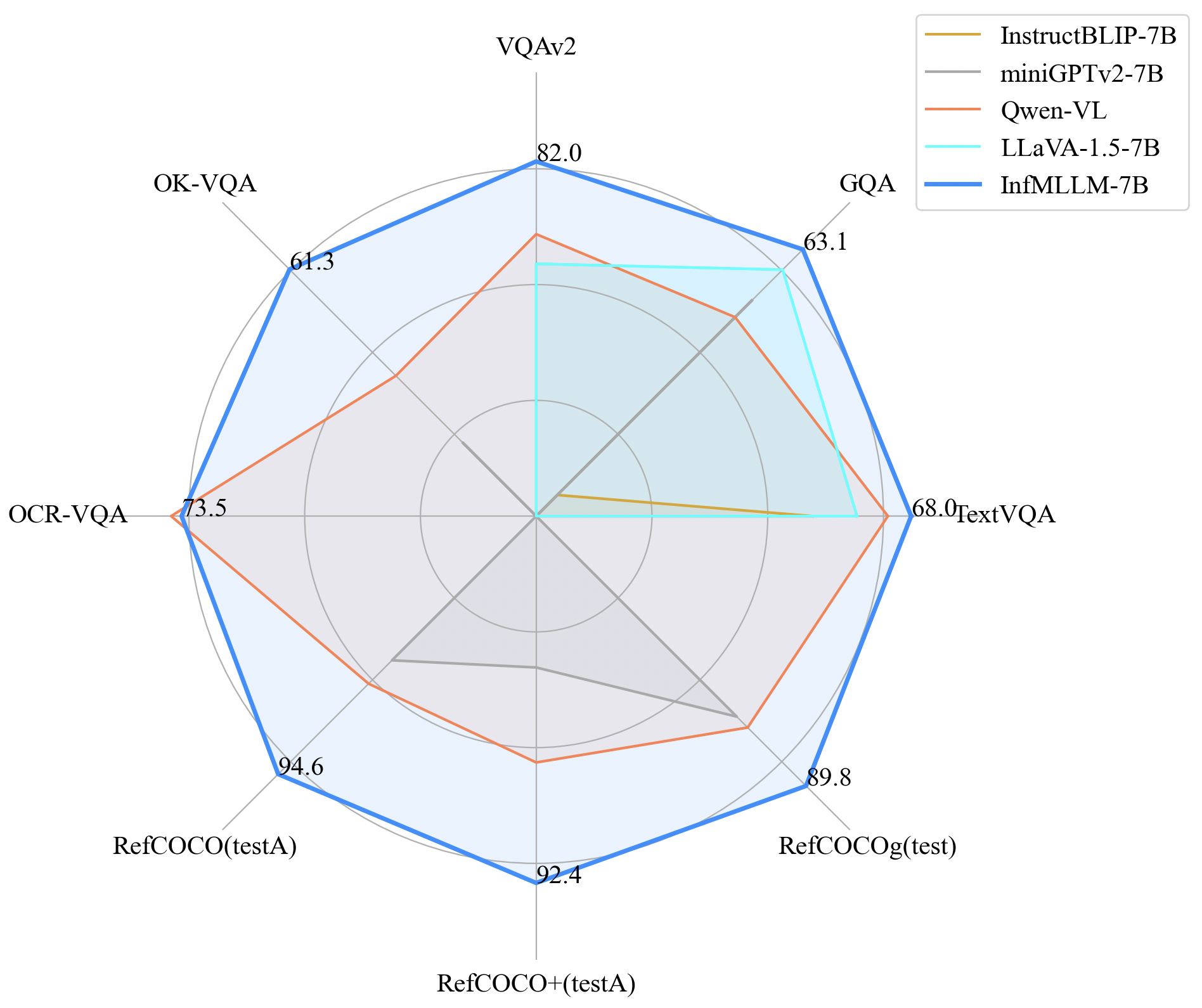}
         \caption{InfMLLM-7B}
    \end{subfigure}
    \begin{subfigure}[b]{0.92\linewidth}
         \centering
         \includegraphics[width=0.92\linewidth]{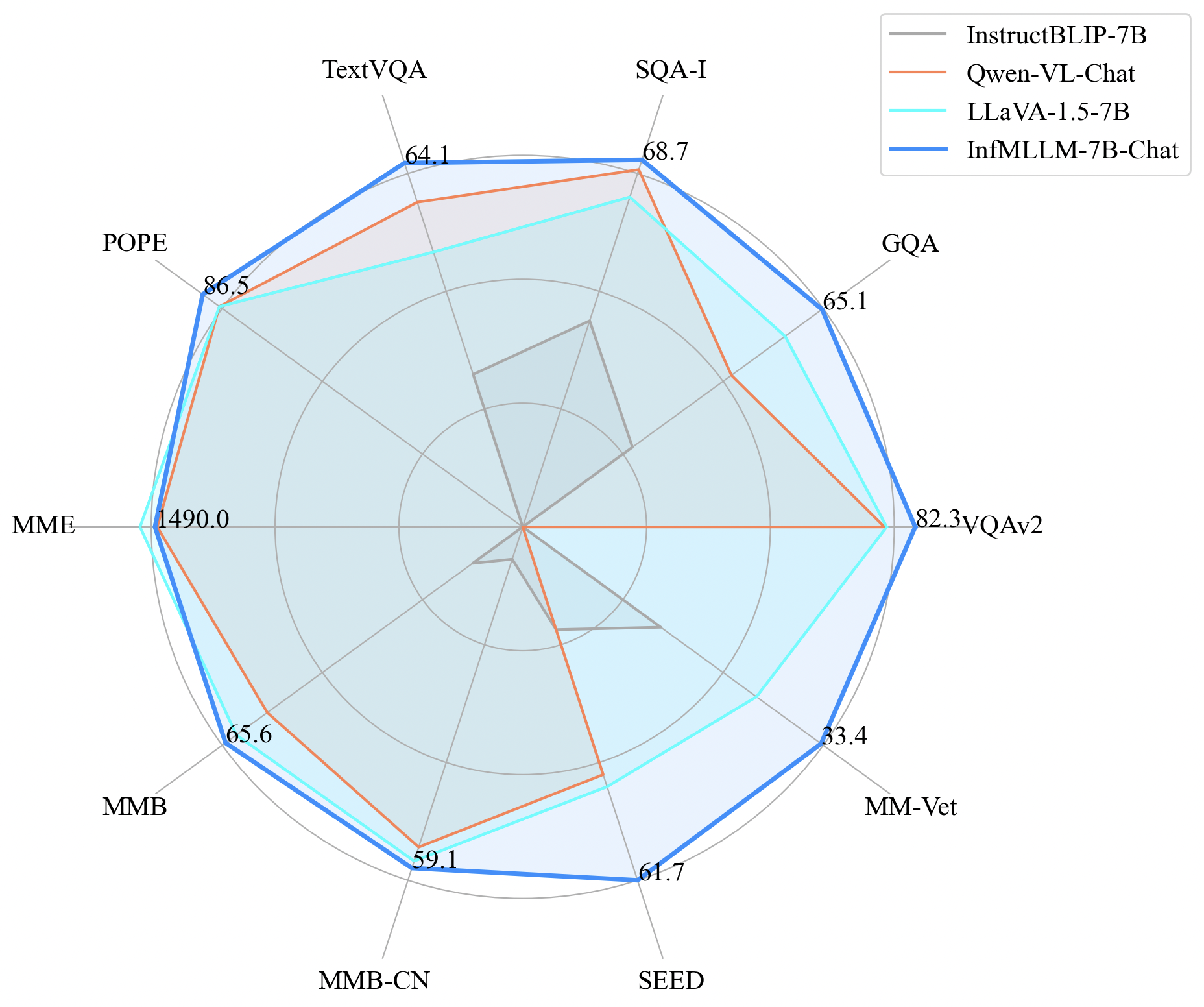}
         \caption{InfMLLM-7B-Chat}
    \end{subfigure}
    \caption{InfMLLM achieves state-of-the-art or similar performance in various vision-language tasks.}
\end{figure}


Effectively finetuning LLMs to extend their capabilities to multimodality remains an ongoing research challenge. This work conducts a progressive training strategy to imporve the training efficiency. In the initial stage, we employ an extensive collection of lower-quality image-text pair data to train a randomly initialized visual adapter. This adapter aligns image features extracted from the pretrained image encoder with text embeddings, allowing the pretrained LLM to process them effectively. 
During the second phase, we utilize publicly available high-quality datasets spanning various vision-language tasks and finetune the model to extend its ability to handle various tasks. To minimize the training overhead during this stage, we freeze the parameters of the LLM except for the QV projection weights. 
To further improve the model's ability to follow instructions effectively, we finetune the entire LLM using a limited amount of instruction data in the final stage.


Visual adapters are commonly used to transform visual features into aligned visual embeddings, which can then be processed by a pretrained LLM. They also help to reduce the number of visual embeddings, which is especially important for large images. Two well-known visual adapters are Q-Former~\cite{li2023blip2} and Perceiver~\cite{alayrac2022flamingo}, which use cross-attention modules to condense a large set of visual features into a small number of visual embeddings.
While these adapters are good at multimodal tasks like image captioning and Visual Question Answering (VQA), they struggle with tasks that involve spatial relationships, such as visual grounding.
In this work, we propose a new visual adapter called pool-adapter. It pools image features into a fixed number of visual features and then converts them into visual embeddings using a two-layer MLP.
Pool-adapter is simple, can be used with different image sizes, and effectively retains the positional information of visual embeddings. This makes it particularly effective for tasks such as visual grounding.

In summary, our main contributions are:
\begin{itemize}
    \item We present a new MLLM framework named InfMLLM that trains in three stages, each with a distinct focus: pool-adapter alignment pretraining, multitask finetuning, and instruction tuning.
    \item We introduce a straightforward pool-adapter to align ViT image features with LLMs, preserving positional information while reducing the number of image embeddings.
    \item We comprehensively evaluated InfMLLM on various benchmarks, achieving either state-of-the-art (SOTA) performance or performance comparable to recent MLLMs.
\end{itemize}

\section{Related Work}
In this section, we survey related research on large language models (LLMs) and multimodal large language models (MLLMs).

\paragraph{Large Language Models (LMMs).}
The landscape of LLMs has evolved significantly, with several noteworthy contributions shaping the field. Early models, such as GPT-2~\cite{radford2019language_GPT2} and BERT~\cite{devlin2019bert}, laid the foundation by demonstrating the potential of training on vast web-scale text datasets. These models marked a breakthrough in Natural Language Processing (NLP) and set the stage for subsequent advancements.
One prominent advancement is GPT-3~\cite{brown2020language_GPT3}, an unprecedented scale and complexity model. GPT-3 showcased the capabilities of massive neural networks, boasting 175 billion parameters and excelling in diverse language tasks. Its release has spurred interest in exploring model size limits and has opened new avenues for understanding the potential applications and challenges associated with such colossal language models.
Megatron-turing NLG~\cite{smith2022using_megatron}, PaLM~\cite{chowdhery2022palm}, Gopher~\cite{rae2022scaling_gopher}, Chinchilla~\cite{hoffmann2022training_chinchilla}, OPT~\cite{zhang2022opt}, and BLOOM~\cite{workshop2023bloom} represent subsequent models that have further pushed the boundaries of LLMs. These models vary in architecture, training methodologies, and applications, contributing to a rich tapestry of research in the domain of large language models. The diversity of approaches reflects ongoing efforts to optimize performance, efficiency, and generalization across different linguistic tasks.
Recent endeavors in the field have concentrated on refining LLMs to better align with human instructions and feedback. InstructGPT~\cite{ouyang2022training_instructGPT}, ChatGPT~\cite{chatgpt} and GPT4~\cite{openai2023gpt4} stand out as exemplars in this regard. These models can engage in dynamic and contextually rich conversations, respond adeptly to user prompts, and even demonstrate proficiency in complex tasks such as code generation. The emphasis on aligning LLMs with human interaction and instruction represents a crucial step toward their practical deployment and integration into real-world applications.

In this work, we investigate a model that is trained end-to-end, with a LLM at its core, aiming to enhance its capabilities in vision-language tasks.

\paragraph{Multimodal Large Language Models (MLLMs).}
Utilizing the impressive generalization capabilities ingrained in LMMs, researchers have undertaken comprehensive investigations to extend their functionalities into multimodal domains. As illustrated by VisualGPT~\cite{Chen_2022_visualgpt} and Frozen~\cite{tsimpoukelli2021multimodal_frozen}, initial undertakings train visual encoders to represent each image through continuous embeddings. These image embeddings were subsequently fed into pretrained language models, enhancing visual language capabilities and focusing on tasks like image captioning and visual question answering. This seminal exploration laid the foundation for subsequent advancements in visual language research, as exemplified by notable studies such as Flamingo~\cite{alayrac2022flamingo}, BLIP-2~\cite{li2023blip2} and Kosmos-1~\cite{huang2023language_kosmos}.
Recent studies, such as VisionLLM~\cite{wang2023visionllm}, Kosmos-2~\cite{peng2023kosmos2}, RegionBLIP~\cite{zhou2023regionblip}, Shikra~\cite{chen2023shikra}, Qwen-VL~\cite{Qwen-VL}, and miniGPT-v2~\cite{chen2023minigptv2}, are actively broadening the spectrum of vision-language tasks that MLLMs can adeptly tackle. These investigations underscore that MLLMs excel in conventional tasks like image captioning and VQA, and exhibit robust visual grounding capability. Specifically, these models generate textual representations of bounding boxes through language models.
Beyond diversifying the tasks within the purview of MLLMs, there is a considerable body of ongoing research in the realm of multimodal instruction tuning. Notable studies in this domain include LLaVA~\cite{liu2023visual_llava}, InstructBLIP~\cite{instructblip}, Otter~\cite{li2023otter}, LLaVA-1.5~\cite{llava_v15_liu2023improved}, and others.

The design of the model architecture and training pipeline of MLLMs is still an open problem. In this work, we propose an efficient model named InfMLLM, which performs appreciatively on various vision-language tasks.
\section{Method}

\begin{figure*}[!t]
    \centering
    \includegraphics[width=1.0\linewidth]{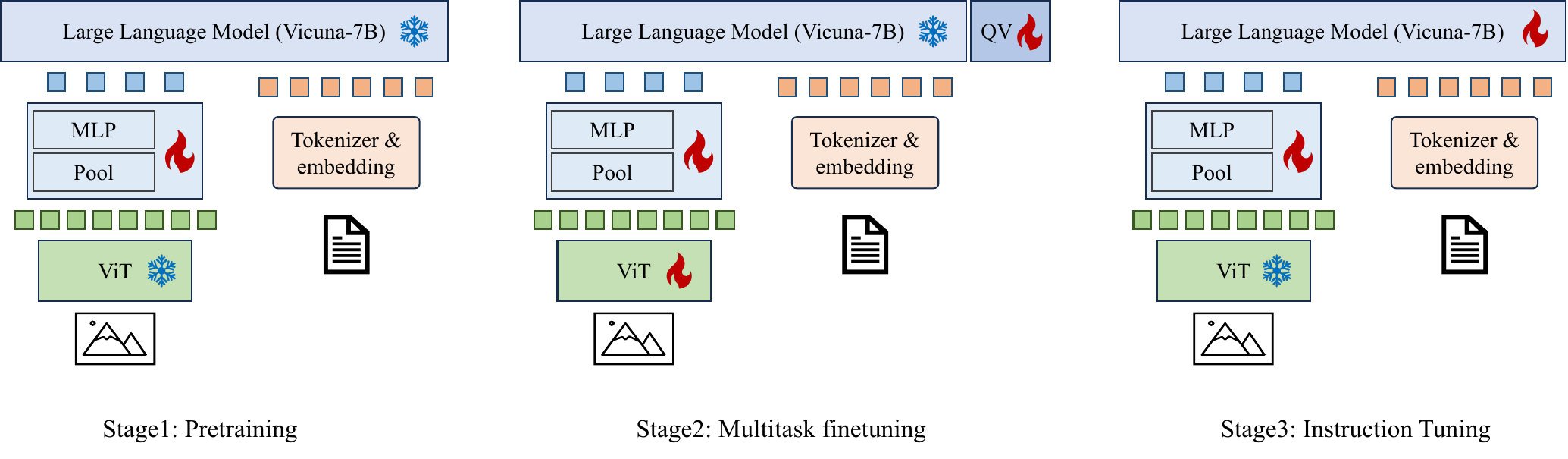}
    \caption{The framework and training pipeline of \ours{}.
    InfMLLM consists of three modules: ViT~\cite{dosovitskiy2020vit} model, pool-adapter, and large language model (LLM). To extend the tasks processed by InfMLLM and improve the instruction-following capability, InfMLLM adopts a three-stage training pipeline.}
    \label{fig:structure_and_pipeline}
\end{figure*}

This section presents a comprehensive explanation of the architecture~\ref{sec:structrue} and training pipeline~\ref{sec:training} of \ours{}.

\subsection{Model Architecture}
\label{sec:structrue}
As depicted in Figure~\ref{fig:structure_and_pipeline}, \ours{} primarily comprises three components: the image encoder, the pool-adapter, and the large language model (LLM).

\paragraph{Image encoder.} 
The image encoder is responsible for the transformation of raw image input into compact and informative image features.
The Vision Transformer~\cite{dosovitskiy2020vit} (ViT) model's streamlined architecture makes it particularly adept at handling diverse multimodal data types, such as images and videos.
Pioneering work like CLIP~\cite{radford2021_clip} further popularized ViT models by providing pretrained models for image-to-text matching. As a result, ViT has become the de facto image encoder of choice in many MLLMs~\cite{liu2023llava, llava_v15_liu2023improved,Qwen-VL,chen2023minigptv2}.
In this work, we employ ViT-g/14 from EVA-CLIP~\cite{EVA-CLIP}. Following the practices in BLIP-2~\cite{li2023blip2}, we discard the final layer of the ViT and utilize the output features of the second last layer. Furthermore, we remove the output features associated with the class token and retain only the patch features.

\paragraph{Pool-adapter.}
LLMs are typically pretrained using pure text corpora and lack the inherent capability to process image features obtained from ViT models.
To bridge the gap between image features and text embeddings, the Q-Former module was introduced in BLIP-2~\cite{li2023blip2}. It serves as a pivotal intermediary, facilitating the connection between ViT and LLMs in the realm of visual understanding. Nonetheless, our initial experiments exposed shortcomings in the performance of Q-Former in tasks related to visual grounding. This discrepancy may be attributed to the tendency of Q-Former's queried features to lose positional information inherent in image features, a critical element for tasks such as visual grounding.
To maximize the retention of positional information in image features, we propose the lightweight pool-adapter. 
Pool-adapter operates by initially pooling the image features, reducing them to a fixed number. Subsequently, a two-layer Multilayer Perceptron (MLP) is employed to effectively align these pooled features with text embeddings, enhancing the model's capacity for multimodal tasks.
Our ablation results in Table~\ref{tbl:ablation_p_multitask} and Table~\ref{tbl:ablation_p_chat} reveal that the number of visual embeddings generated by the pool-adapter substantially influences the model's performance on vision-language tasks. Methods of maintaining a reduced number of visual embeddings while enhancing model performance warrant further investigation, as a reduction in visual embeddings can considerably improve the model inference speed.

\paragraph{LLM.}
The LLM plays a central role in multimodal large language models. They receive instructions as well as aligned image features as input and then generate corresponding answers.
In this work, we use the open source Vicuna-7B~\cite{vicuna2023}, a powerful large language model.
Vicuna was initially finetuned using pure text corpora, and further finetuning it with multimodal data will benefit the performance of MLLMs on multimodal tasks. For example, Qwen-VL~\cite{Qwen-VL} finetunes the full LLM in the multitask pretraining phase, and achieves excellent performance on tasks such as captioning, visual question answering (VQA), text-oriented VQA, and visual grounding.
As LLMs get larger, the full finetuning approach becomes less feasible. LoRA~\cite{hu2021lora} finetuning is a more efficient approach and is adopted in miniGPT-v2~\cite{chen2023minigptv2}.
In our initial experiments, the LoRA finetuned model performed at a disadvantage on vision-language tasks compared to the fully finetuned model. 
To achieve an optimal balance between performance and training efficiency, we opt to finetune the complete QV projection weights within the LLM while maintaining the remaining parameters frozen, as illustrated in the multitask finetuning stage in Figure~\ref{fig:structure_and_pipeline}.
Note that during the final instruction tuning stage, we freeze the ViT model and finetune the entire LLM model.

\begin{table*}[!t]
\centering
\resizebox{0.95\linewidth}{!}{ 
\begin{tabular}{@{}l|ll@{}}
\toprule
                         & \multicolumn{2}{c}{Prompt}          \\ \midrule
Stage 1                  & \multicolumn{2}{l}{\textless{}image\textgreater{}\textless{}ImageHere\textgreater{}A short image caption: \textcolor{red}{A child holding a flowered umbrella and petting a yak.}}                                                                                                   \\ \midrule
\multirow{3}{*}{Stage 2} & \multicolumn{1}{l|}{VQA}        & \textless{}image\textgreater{}\textless{}ImageHere\textgreater{}Question: What color is the bedspread? Short answer: \textcolor{red}{white}                                                                                                        \\
                         & \multicolumn{1}{l|}{Captioning} & \textless{}image\textgreater{}\textless{}ImageHere\textgreater{}A short image caption: \textcolor{red}{A bath tub sitting next to a sink in a bathroom.}  \\
                         & \multicolumn{1}{l|}{Grounding}  & \textless{}image\textgreater{}\textless{}ImageHere\textgreater{}\textless{}ref\textgreater{}woman in top picture on the left\textless{}/ref\textgreater{}\textcolor{red}{\textless{}box\textgreater{}(0.074,0.142),(0.390,0.468)\textless{}/box\textgreater{}} \\ \midrule
Stage 3                  & \multicolumn{2}{l}{LLaVA-1.5~\cite{llava_v15_liu2023improved}}  \\ \bottomrule
\end{tabular}
}
\caption{Prompts employed in \ours{}. It's worth noting that ``\textless{}ImageHere\textgreater{}" is substituted with the aligned visual embeddings from the pool-adapter, and the text in red represents the ground-truth label.}
\label{tbl:prompts}
\end{table*}

\begin{table}[!t]
\centering
\resizebox{1.0\linewidth}{!}{ 
\begin{tabular}{@{}l|l@{}}
\toprule
                         & \multicolumn{1}{c}{Dataset}                                                        \\ \midrule
Stage 1                  & CC3M, CC12M, LAION-115M                   \\ \midrule
\multirow{3}{*}{Stage 2} & VQA: VQAv2, OK-VQA, AOK-VQA, GQA, OCR-VQA, TextVQA \\ 
                         & Captioning: COCO,TextCaps                                  \\ 
                         & Grounding: RefCOCO, RefCOCO+, RefCOCOg                     \\ \midrule
Stage 3                  & LLaVA\_1.5\_665k                                               \\ \bottomrule
\end{tabular}
}
\caption{Training dataset used in stage-1 pretraining and stage-2 multitask finetuning.}
\label{tbl:training_data}
\end{table}

\subsection{Data, Prompts and Training}
\label{sec:training}
In this subsection, we provide the datasets, prompts, and training details for each stage of \ours{}.

\paragraph{Stage-1 Pretraining.}
In this phase, we employ weakly labeled image-text pairs to conduct pretraining of the pool-adapter module. The pretrained pool-adapter will align the image features derived from ViT to text embeddings, enabling recognition by pretrained LLMs.
Throughout this stage, all parameters are freezed, except for those of pool-adapter. The data utilized for pretraining consist of publicly available datasets, including CC3M, CC12M, and LAION-115M~\cite{li2022blip}.
During this phase, we employ a straightforward prompt, where LLMs are prompted to generate a concise caption based on the aligned image features, as exemplified in Table~\ref{tbl:prompts}.

\paragraph{Stage-2 Multitask finetuning.}
By aligning image features to text embeddings that are compatible with LLM processing, various visual language tasks such as image captioning, visual question answering (VQA), and referring expression grounding can be integrated into a unified MLLM model. 
In this work, we collect publicly available datasets, encompassing image captioning datasets (COCO~\cite{COCO_LinMBHPRDZ14}, TextCaps~\cite{sidorov2019textcaps}), VQA datasets (VQAv2~\cite{vqav2_GoyalKSBP17}, OK-VQA~\cite{okvqa_MarinoRFM19}, AOK-VQA~\cite{aokvqa_SchwenkKCMM22}, GQA~\cite{gqa_HudsonM19}, OCR-VQA~\cite{ocrvqa_0001SSC19}, TextVQA~\cite{sidorov2019textcaps}), and referring expression grounding datasets (RefCOCO~\cite{refcoco_KazemzadehOMB14}, RefCOCO+, RefCOCOg), and subsequently finetune a general MLLM model.
In this phase, we conduct finetuning on the ViT model, the pool-adapter, and the QV projection weights within the LLM.
We employ uniform sampling of training data from three tasks: image captioning, visual question answering, and visual grounding. The prompts utilized for the data associated with these three tasks exhibit slight variations, as outlined in Table~\ref{tbl:prompts}. Specifically, we employ identical prompts for VQA and text-oriented VQA, without making any distinctions.
For visual grounidng tasks, we utilize the LLM to directly generate the normalized coordinates of the upper left and lower right corners of the bounding box.

\paragraph{Stage-3 Instruction Tuning.}

Instruction tuning~\cite{wei2022finetuned_flan} is frequently a crucial step in improving LMMs' zero-shot performance and ability to follow instructions, and its effectiveness has been well-established.
In order to improve InfMLLM's instruction-following capability, we further finetune the model with instruction data.
Recent efforts~\cite{instructblip,liu2023llava,llava_v15_liu2023improved,chen2023sharegpt4v}, whether done manually or leveraging GPT-4, involve the transformation of public datasets into usable instruction tuning datasets.
In this work, we employ the instructional data provided by LLaVA-1.5~\cite{llava_v15_liu2023improved}.
%
%
During the instruction tuning stage, all parameters, except those of ViT, undergo the finetuning process.

\begin{figure*}[!t]
    \centering
    \includegraphics[width=1.0\linewidth]{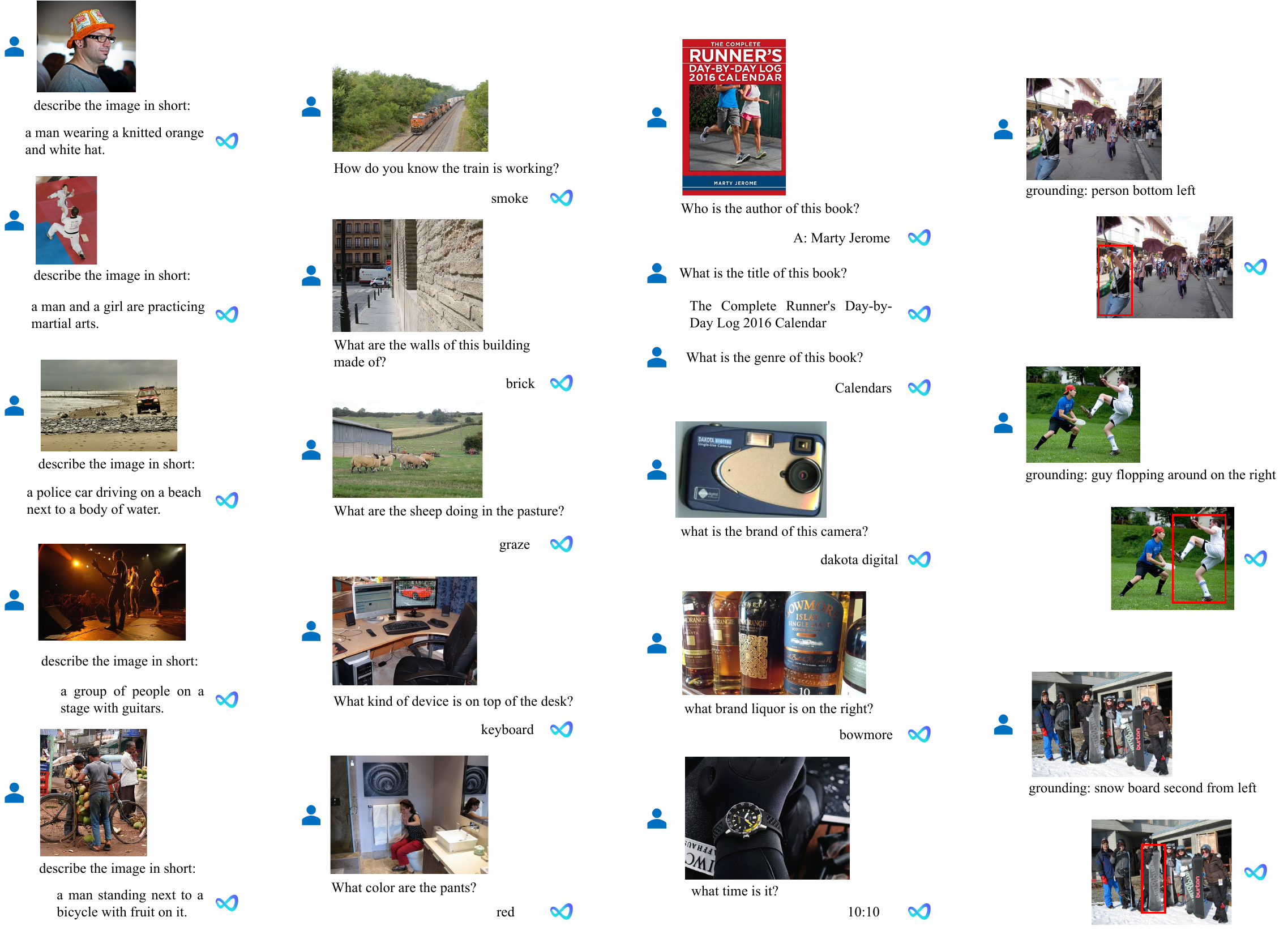}
    \caption{Some qualitative examples generated by InfMLLM-7B. The capabilities of \ours{} encompass support for image captioning, visual question answering (VQA), text-oriented VQA, and visual grounding.}
    \label{fig:demo_multitask}
\end{figure*}

\begin{table*}[!t]
\centering
\resizebox{1.0\linewidth}{!}{
\begin{tabular}{@{}l|l|cccccccc@{}}
\toprule
\multirow{2}{*}{Model} & \multirow{2}{*}{LLM} & \multicolumn{8}{c}{Benchamrks}          \\ \cmidrule(l){3-10}

                       &     & \multicolumn{1}{c}{VQAv2} &\multicolumn{1}{c}{OK-VQA} & \multicolumn{1}{c}{GQA} & \multicolumn{1}{c}{TextVQA} & \multicolumn{1}{c|}{OCR-VQA} & \multicolumn{1}{c|}{\begin{tabular}[c]{@{}c@{}}RefCOCO\\ val / test-A / test-B\end{tabular}} & \multicolumn{1}{c|}{\begin{tabular}[c]{@{}c@{}}RefCOCO+\\ val / test-A / test-B\end{tabular}} & \begin{tabular}[c]{@{}c@{}}RefCOCOg\\ val / test\end{tabular} \\ 
                       \midrule  
                       
            Flamingo-9B~\cite{alayrac2022flamingo}                &    & \multicolumn{1}{c}{-}         & \multicolumn{1}{c}{44.7}       & \multicolumn{1}{c}{-}    & \multicolumn{1}{c}{-}        & \multicolumn{1}{c|}{-}        & \multicolumn{1}{c|}{-}                                                                       & \multicolumn{1}{c|}{-}              &           -    \\
            
            BLIP-2~\cite{li2023blip2}              &    Vicuna-13B         & \multicolumn{1}{c}{-}   & \multicolumn{1}{c}{45.9}       & \multicolumn{1}{c}{32.3}    & \multicolumn{1}{c}{42.4}        & \multicolumn{1}{c|}{-}        & \multicolumn{1}{c|}{-}      & \multicolumn{1}{c|}{-}       &       -            \\   
            
            InstructBLIP~\cite{instructblip}              &    Vicuna-13B    &    \multicolumn{1}{c}{-}     & \multicolumn{1}{c}{-}       & \multicolumn{1}{c}{45.9}    & \multicolumn{1}{c}{50.7}        & \multicolumn{1}{c|}{-}        & \multicolumn{1}{c|}{-}                                                                       & \multicolumn{1}{c|}{-}         &        -      \\

        OFA-L~\cite{wang2022ofa}             &        & \multicolumn{1}{c}{-}          & \multicolumn{1}{c}{-}       & \multicolumn{1}{c}{-}    & \multicolumn{1}{c}{-}        & \multicolumn{1}{c|}{-}        & \multicolumn{1}{c|}{79.96 / 83.67 / 76.39}                                                                       & \multicolumn{1}{c|}{68.29 / 76.00 / 61.75}        &   67.57 / 67.58         \\ 

        VisionLLM-H~\cite{wang2023visionllm}            &        & \multicolumn{1}{c}{-}          & \multicolumn{1}{c}{-}       & \multicolumn{1}{c}{-}    & \multicolumn{1}{c}{-}        & \multicolumn{1}{c|}{-}        & \multicolumn{1}{c|}{- / 86.70 / -}                                                                       & \multicolumn{1}{c|}{-}        &   -        \\

        mPLUG-DocOwl~\cite{ye2023mplugdocowl}          &   LLaMA-7B     & \multicolumn{1}{c}{-}          & \multicolumn{1}{c}{-}       & \multicolumn{1}{c}{-}    & \multicolumn{1}{c}{52.6}        & \multicolumn{1}{c|}{-}        & \multicolumn{1}{c|}{-}                                                                       & \multicolumn{1}{c|}{-}        &   -        \\ 

        Shikra-7B~\cite{chen2023shikra}             &   Vicuna-7B    & \multicolumn{1}{c}{-}          & \multicolumn{1}{c}{-}       & \multicolumn{1}{c}{-}    & \multicolumn{1}{c}{-}        & \multicolumn{1}{c|}{-}        & \multicolumn{1}{c|}{87.01 / 90.61 / 80.24}                                                                       & \multicolumn{1}{c|}{81.60 / 87.36 / 72.12}        &   82.27 / 82.19         \\ 
        
        Shikra-13B~\cite{chen2023shikra}          &   Vicuna-13B          & \multicolumn{1}{c}{-}   & \multicolumn{1}{c}{47.16}       & \multicolumn{1}{c}{-}    & \multicolumn{1}{c}{-}        & \multicolumn{1}{c|}{-}        & \multicolumn{1}{c|}{87.83 / 91.11 / 81.81}                                                                       & \multicolumn{1}{c|}{82.89 / 87.79 / 74.41}        &   82.64 / 83.16         \\

      MiniGPT-v2-7B~\cite{chen2023minigptv2}      &   LLaMA2-7B    &  \multicolumn{1}{c}{-}          & \multicolumn{1}{c}{56.9}       & \multicolumn{1}{c}{60.3}    & \multicolumn{1}{c}{-}        & \multicolumn{1}{c|}{-}        & \multicolumn{1}{c|}{88.69 / 91.65 / 85.33}                                    & \multicolumn{1}{c|}{79.97 / 85.12 / 74.45}        &     84.44 / 84.66       \\

        QWen-VL-7B~\cite{Qwen-VL}      &   Qwen-7B    & \multicolumn{1}{c}{79.5}          & \multicolumn{1}{c}{58.6}       & \multicolumn{1}{c}{59.3}    & \multicolumn{1}{c}{63.8}        & \multicolumn{1}{c|}{\textbf{75.7}}        & \multicolumn{1}{c|}{89.36 / 92.26 / 85.34}                                                                       & \multicolumn{1}{c|}{83.12 / 88.25 / 77.21}        &      85.58 / 85.48      \\

                LLaVA-1.5-7B~\cite{llava_v15_liu2023improved}       &   Vicuna-7B             & \multicolumn{1}{c}{78.5} & \multicolumn{1}{c}{-}       & \multicolumn{1}{c}{62.0}    & \multicolumn{1}{c}{58.2}        & \multicolumn{1}{c|}{-}        & \multicolumn{1}{c|}{}                                                                       & \multicolumn{1}{c|}{}        &            \\ 
                \midrule

                 InfMLLM-7B    &   Vicuna-7B     & \multicolumn{1}{c}{\textbf{81.95}}       & \multicolumn{1}{c}{ \textbf{61.33} }       & \multicolumn{1}{c}{ \textbf{63.15} }    & \multicolumn{1}{c}{\textbf{68.02}}        & \multicolumn{1}{c|}{\underline{73.52}}        & \multicolumn{1}{c|}{\;\; - \;\; / \textbf{94.62} / \textbf{89.30}}                                                                       & \multicolumn{1}{c|}{\;\; - \;\; / \textbf{92.36} / \textbf{81.83} }        &   \;\; - \;\; / \textbf{89.78}         \\ 
                     
                    \bottomrule
\end{tabular}
}
\caption{Comparison with SoTA MLLMs across tasks such as visual question answering (VQA), text-oriented VQA, and visual grounding.
InfMLLM-7B achieves the best performance on most benchmarks, and ranks the second on the other.
. Benchmark names are abbreviated due to space limits. VQAv2~\cite{vqav2_GoyalKSBP17}; OK-VQA~\cite{okvqa_MarinoRFM19}; GQA~\cite{gqa_HudsonM19}; TextVQA~\cite{textvqa_SinghNSJCBPR19}; OCR-VQA~\cite{ocrvqa_0001SSC19}; RefCOCO~\cite{refcoco_KazemzadehOMB14}; RefCOCO+~\cite{refcoco_KazemzadehOMB14}; RefCOCOg~\cite{refcoco_KazemzadehOMB14}.
}
\label{tbl:results_multitask}
\end{table*}

\begin{figure*}[!t]
    \centering
    \includegraphics[width=0.88\linewidth]{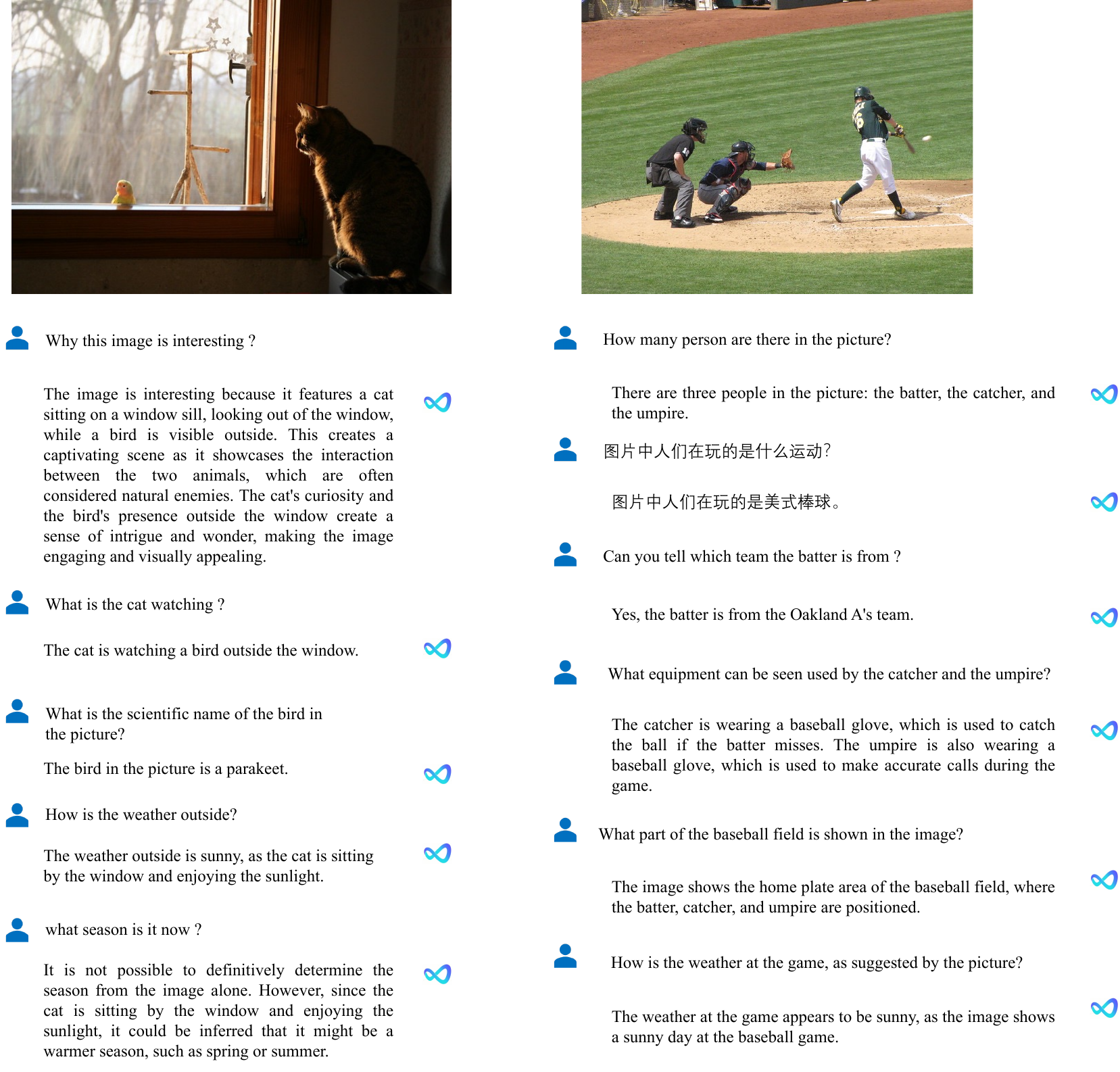}
    \caption{Some qualitative examples generated by InfMLLM-7B-Chat.}
    \label{fig:demo_chat}
\end{figure*}

\begin{table*}[!t]
\centering
\resizebox{0.88\linewidth}{!}{

\begin{tabular}{@{}l|l|cccc|cccccc@{}}
\toprule
Method          & LLM       & VQAv2 & GQA   & SQA$^{\text{I}}$ & VQA$^{\text{T}}$ & POPE  & MME  & MMB   & MMB$^{\text{CN}}$ & SEED  & MM-Vet \\ \midrule

BLIP-2 & Vicuna-13B & 41.0 & 41 &  61 & 42.5 & 85.3 & 1294 & – & – &46.4 & 22.4\\

InstructBLIP & Vicuna-7B & - & 49.2 & 60.5 & 50.1 & - & - & 36 & 23.7 & 53.4 & 26.2 \\
InstructBLIP & Vicuna-13B & - & 49.5 & 63.1 & 50.7 & 78.9 & 1212 & - & - &  - & 25.6 \\

Shikra & Vicuna-13B & 77.4 & - & - & - & - & - & 58.8 & - & - & - \\

IDEFICS-9B & LLaMA-7B & 50.9 & 38.4 &  –  & 25.9 & - & - & 48.2 & 25.2 & - & - \\
IDEFICS-80B &  LLaMA-65B & 60.0 & 45.2  & –  &30.9 &– &– & 54.5 & 38.1 & - & - \\

Qwen-VL-Chat & Qwen-7B & 78.2 & 57.5 & 68.2 & 61.5 &  - &  1487 & 60.6 & 56.7 & 58.2 & - \\

LLaVA-1.5-7B    & Vicuna-7B &   78.5    &  62.0     &   66.8    & 58.2                       &   85.9    &   \textbf{1510}   &   64.3    &  58.3       &   58.6    &  30.5      \\ \midrule

InfMLLM-7B-Chat & Vicuna-7B &    \textbf{82.3}   & \textbf{65.1} & \textbf{68.7}       & \textbf{64.1}    & \textbf{86.5} & \underline{1490} & \textbf{65.6} &  \textbf{59.1}  & \textbf{61.7} &  \textbf{33.4} \\ 

\bottomrule
\end{tabular}

}
\caption{
Comparison with SoTA methods on 11 benchmarks. InfMLLM-7B-Chat achieves the best performance on most benchmarks, and ranks the second on the other.
Benchmark names are abbreviated due to space limits. VQAv2~\cite{vqav2_GoyalKSBP17}; GQA~\cite{gqa_HudsonM19}; SQAI: ScienceQA-IMG~\cite{lu2022sqa};
VQAT: TextVQA~\cite{textvqa_SinghNSJCBPR19}; POPE~\cite{pope_2305}; MME~\cite{fu2023mme}; MMB: MMBench~\cite{liu2023mmbench}; MMBCN: MMBench-Chinese~\cite{liu2023mmbench}; SEED: SEED-Bench~\cite{seedbench-2307-16125}; MM-Vet~\cite{mmvet-2308-02490}.
}
\label{tbl:results_mme}
\end{table*}

\begin{table*}[!t]
\centering
\resizebox{0.9 \linewidth}{!}{ 
\begin{tabular}{@{}c|cccccccc|c@{}}
\toprule
\multirow{2}{*}{$p$} & \multicolumn{8}{c|}{Benchmarks}                                                                                                                                                                                                                                            & \multirow{2}{*}{Mean ($
\uparrow$)} \\ \cmidrule(lr){2-9}

    & \multicolumn{1}{c}{OKVQA} & \multicolumn{1}{c}{GQA}   & \multicolumn{1}{c}{VQAv2} & \multicolumn{1}{c}{TextVQA} & \multicolumn{1}{c}{OCR-VQA} & \multicolumn{1}{c}{RefCOCO testA} & \multicolumn{1}{c}{RefCOCO+ testA} & RefCOCOg test &                      \\ \midrule
                            
 8   & \multicolumn{1}{c}{59.51} & \multicolumn{1}{c}{61.70} & \multicolumn{1}{c}{78.75}      & \multicolumn{1}{c}{63.78}   & \multicolumn{1}{c}{70.81}         & \multicolumn{1}{c}{92.59}         & \multicolumn{1}{c}{88.71}          & 86.04         &         75.23             \\

 16                               & \multicolumn{1}{c}{59.22} & \multicolumn{1}{c}{61.75} & \multicolumn{1}{c}{79.85}      & \multicolumn{1}{c}{62.19}   & \multicolumn{1}{c}{71.26}         & \multicolumn{1}{c}{93.49}         & \multicolumn{1}{c}{89.76}          & 87.20         &          75.59            \\
                            
32                               & \multicolumn{1}{c}{60.26} & \multicolumn{1}{c}{63.53} & \multicolumn{1}{c}{81.54}      & \multicolumn{1}{c}{66.89}   & \multicolumn{1}{c}{72.46}    & \multicolumn{1}{c}{94.87}         & \multicolumn{1}{c}{92.35}          & 89.50         & 77.67                \\

8+16+32                          & \multicolumn{1}{c}{\textbf{61.33}} & \multicolumn{1}{c}{\underline{63.15}} & \multicolumn{1}{c}{\textbf{81.95}}      & \multicolumn{1}{c}{\textbf{68.02}}   & \multicolumn{1}{c}{\textbf{73.52}}    & \multicolumn{1}{c}{\underline{94.62}}         & \multicolumn{1}{c}{\textbf{92.36}}          & \textbf{89.78}         & \textbf{78.09}                \\ 
                            \bottomrule
\end{tabular}
}
\caption{Ablation studies on the configuration of parameter $p$ in pool-adapter. The model is InfMLLM-7B after stage-2 multitask finetuning.
}
\label{tbl:ablation_p_multitask}
\end{table*}

\begin{table*}[!t]
\centering
\resizebox{1.0\linewidth}{!}{ 
\begin{tabular}{@{}c|cccc|cccccc@{}}
\toprule
 $p$       & VQAv2 & GQA   & SQA$^{\text{I}}$ & VQA$^{\text{T}}$ & POPE  & MME  & MMB   & MMB$^{\text{CN}}$ & SEED  & MM-Vet \\ \midrule
 
16 & 80.16 $\pm$ 0.02 & 63.69 $\pm$ 0.01 & 62.42 $\pm$ 0.39 & 61.58 $\pm$ 0.21 & 84.80 $\pm$ 0.08 & 1434 $\pm$ 9 & 64.45 $\pm$ 0.17 & 54.46 $\pm$ 0.01 & 59.36 $\pm$ 0.08 & 32.90 $\pm$ 1.50 \\ 
        
32 & 81.50 $\pm$ 0.17 & \textbf{64.36} $\pm$ 0.40 & \textbf{66.63} $\pm$ 0.99 & 63.57 $\pm$ 1.72 & \textbf{86.71} $\pm$ 1.13 & \textbf{1494} $\pm$ 12 & 65.21 $\pm$ 0.25 & 56.86 $\pm$ 1.11 & 61.26 $\pm$ 0.14 & 33.35 $\pm$ 0.05 \\
        
8+16+32 & \textbf{81.99} $\pm$ 0.08 & \underline{64.13} $\pm$ 0.16 & \underline{66.01} $\pm$ 0.86 & \textbf{65.26} $\pm$ 0.14 & \underline{85.86} $\pm$ 0.30 & 1476 $\pm$ 7 & \textbf{65.73} $\pm$ 0.09 & \textbf{57.94} $\pm$ 0.30 & \textbf{61.77} $\pm$ 0.02 & \textbf{34.65} $\pm$ 0.05 \\

\bottomrule
\end{tabular}
}
\caption{Ablation studies on the configuration of parameter $p$ in pool-adapter. The model is InfMLLM-7B-Chat after stage-3 instruction tuning.
The chat model's performance shows considerable variation across certain benchmarks; therefore, to ensure reliability, each experimental setup was executed twice. The average performance and the standard deviation for each configuration are provided for a clear comparison.}
\label{tbl:ablation_p_chat}
\end{table*}

\begin{table}[!t]
\centering
\resizebox{0.65\linewidth}{!}{ 
\begin{tabular}{@{}c|ccc@{}}
\toprule
Hyperparameters & \multicolumn{1}{c}{Stage-1} & \multicolumn{1}{c}{Stage-2} & Stage-3 \\ \midrule
batch size      & \multicolumn{1}{c}{1024}        & \multicolumn{1}{c}{512}        &    128     \\ 
lr              & \multicolumn{1}{c}{2e-4}        & \multicolumn{1}{c}{2e-5}        &    2e-5     \\ 
lr scheduler    & \multicolumn{3}{c}{Cosine decay}                                                  \\
optimizer       & \multicolumn{3}{c}{AdamW}                                                  \\ 
image resolution           & \multicolumn{1}{c}{224}        & \multicolumn{1}{c}{448}        &     448    \\ 
DeepSpeed stage & \multicolumn{1}{c}{2}        & \multicolumn{1}{c}{2}        &       2  \\ \bottomrule
\end{tabular}
}
\caption{Training hyperparameters of \ours{}.}
\label{tbl:hyperparameter}
\end{table}

\begin{table}[!t]
\centering
\resizebox{0.9\linewidth}{!}{
\begin{tabular}{@{}l|l|l|l@{}}
\toprule
Task      & Dataset   & Evaluation split & Metric   \\ \midrule

Image Caption       & Flickr30K & karpathy-test    & CIDEr($\uparrow$)    \\ \midrule

\multirow{3}{*}{General VQA}    & OKVQA     & val              &     VQA Score ($\uparrow$)     \\ 
           & GQA       & test-balanced    &    EM($\uparrow$)      \\
           & VQAv2       & test-dev    &   VQA Score ($\uparrow$)      \\
           \midrule
           
\multirow{2}{*}{Text-oriented VQA}  & TextVQA   & val              &      VQA Score ($\uparrow$)    \\ 
       & OCRVQA    & test             &    EM($\uparrow$)      \\ \midrule
    
\multirow{3}{*}{\begin{tabular}[c]{@{}l@{}}Refer Expression\\ Comprehension\end{tabular}} & RefCOCO   & testA \& testB   & Accuracy($\uparrow$) \\
               & RefCOCO+  & testA \& testB   & Accuracy($\uparrow$) \\ 
             & RefCOCOg  & test             & Accuracy($\uparrow$) \\ \bottomrule
\end{tabular}
}
\caption{Summary of benchmark datasets employed for evaluating the stage-2 multitask finetuning model.}
\label{tbl:eval_datasets}
\end{table}

\section{Experiments}
In this section, we delve into the details of model training and perform a comprehensive quantitative evaluation of the model on public benchmarks. Additionally, we provide a qualitative demonstration with visual examples to provide a more complete understanding of the model's capabilities.

\paragraph{Training hyperparameters.} 
The training hyperparameters are outlined in Table~\ref{tbl:hyperparameter}. In particular, all stage models undergo training using the AdamW optimizer and a cosine decay learning rate scheduler.
For the initial stage-1 pretraining, the model is trained using 32$\times$A800 GPUs with a global batch size of 1024 and a maximum learning rate of 2e-4. The image resolution is set to 224 during this stage.
In the subsequent stage-2 multitask finetuning, the model is trained with 32$\times$A800 GPUs, employing a global batch size of 512 and a maximum learning rate of 2e-5. The image resolution is increased to 448 for this stage.
Lastly, in the instruction tuning stage, the model is trained using 32$\times$A800 GPUs for 1 epoch, with a global batch size of 128 and a maximum learning rate of 2e-5. The image resolution remains at 448 for this stage.

\subsection{Visualization}

To provide a more intuitive illustration of the vision-language tasks accomplished by \ours{}, we have curated visual examples showcased in Figure~\ref{fig:demo_multitask}. The examples reveal \ours{}'s ability to seamlessly integrate tasks such as image captioning, VQA, OCR-related VQA, and visual grounding into a unified model. In one word, InfMLLM effectively extends the task boundaries of LMMs.
Additionally, we offer a selection of multi-turn conversations in Figure~\ref{fig:demo_chat} utilizing the InfMLLM-7B-Chat model to showcase its proficiency in following instructions.

\subsection{Benchmark Evaluation}
Following multitask finetuning, InfMLLM becomes a versatile model capable of addressing various vision and language tasks, such as Visual Question Answering (VQA), text-oriented VQA, and visual grounding.
To comprehensively assess the capabilities of InfMLLM, we undertake evaluations across all these tasks. The specifics regarding the datasets, splits, and evaluation metrics can be found in Table~\ref{tbl:eval_datasets}.
The results presented in Table~\ref{tbl:results_multitask} highlight the strong performance of InfMLLM, achieving state-of-the-art results across nearly all visual grounding and VQA tasks.
%

To assess the instruction-following capability of InfMLLM-Chat after stage-3 finetuning, we evaluated InfMLLM-Chat on various benchmarks. As illustrated in Table~\ref{tbl:results_mme}, InfMLLM-Chat exhibited superior performance compared to other MLLMs employing language models of similar size.

\subsection{Ablation}
In this section, we conduct ablation studies on the InfMLLM-7B and InfMLLM-7B-Chat models.

\paragraph{Effect of number of visual embeddings.}
\label{sec:ablation_number_visual_embeddings}
LLMs possess the ability to extract valuable insights from diverse and complex text sequences.
Similarly, MLLMs are expected to be capable of gathering pertinent information from visual embeddings to effectively answer various questions.
Yet, the optimal number of visual embeddings required for MLLMs remains uncertain. In this section, we explore how the number of visual embeddings affects the performance of InfMLLM-7B and InfMLLM-7B-Chat.
Specifically, we adjust the value of the parameter $p$ in the pool-adapter, which regulates the number of image embeddings (equivalent to $p^2$) fed into the LLMs.
As demonstrated in Table~\ref{tbl:ablation_p_multitask}, there is a clear trend observed: the performance of InfMLLM-7B in visual-language tasks improves with an increase in the number of visual embeddings.
Across eight benchmarks, setting $p$ to 8 yields a mean performance of 75.23. Increasing $p$ to 32 results in a mean performance boost to 77.67. Furthermore, concatenating image embeddings from three settings—8, 16, and 32—further increases the mean performance to 78.09.
The same conclusion is reached in the chat model of InfMLLM-7B-Chat, as shown in Table~\ref{tbl:ablation_p_chat}.

\paragraph{Adjust the quantity of visual embeddings online}

While increasing visual embeddings improves performance, it comes at the cost of slowing down inference. There is a frequent requirement for swift, initial results with fewer visual embeddings. The process of training multiple models for different values of $p$ can be quite burdensome. However, our experiments in Table~\ref{tbl:p_online_setting} demonstrate that when a model is trained with $p=32+16+8$, utilizing $p=32$ or $p=16$ during the inference phase unexpectedly sustains strong performance without substantial degradation.

\begin{table}[!h]
\centering
\resizebox{1.0\linewidth}{!}{ 
\begin{tabular}{@{}c|ccccccc@{}}
\toprule
$p$       & GQA  & SQA-I & TextVQA & POPE & MME  & SEED & MM-Vet \\ \midrule
16      &    61.3  &   66.2    &     55.9    &   82.5   &   1365   &  57.5    &        \\ 
32      & 64.5 & 67.0  & 62.4    & 85.0 & 1444 & 60.9 & 36.1   \\ 
32+16+8 & 65.1 & 68.7  & 64.1    & 86.5 & 1490 & 61.7 & 33.4   \\ \bottomrule
\end{tabular}
}
\caption{
Ablation of the impact of online adjustments to visual embeddings. The model undergoes training with $p$ configured as ``32+16+8'' and is subsequently evaluated with varying $p$ settings during inference.
}
\label{tbl:p_online_setting}
\end{table}

\section{Limitations}
In our experiments, we observed that during the multitask finetuning phase, there exists a degree of optimization conflict in individual tasks. The performance is notably influenced by the choice of loss weights or data distribution ratios for different tasks, necessitating meticulous finetuning. Future research will delve into discovering more effective solutions for multitask finetuning.

\section{Conclusion}
In this work, we introduce a novel MultiModal Large Language Model framework named InfMLLM. Utilizing a straightforward pool-adapter, InfMLLM dynamically adjusts the number of image embeddings while retaining crucial positional information. In contrast to other MLLMs, InfMLLM attains state-of-the-art performance in visual grounding and visual question answering, while also delivering competitive results in image captioning and text-oriented VQA tasks. We will release the associated code and models as open-source. We anticipate that InfMLLM will contribute to the advancement of MLLM-related research.

{
    \small
    \bibliographystyle{ieeenat_fullname}
    \bibliography{main}
}


\end{document}